\documentclass[10pt,twocolumn,letterpaper]{article}

\usepackage{btas}
\usepackage{times}
\usepackage{epsfig}
\usepackage{graphicx}
\usepackage{amsmath}
\usepackage{amssymb}
\usepackage{array}
\newcolumntype{P}[1]{>{\centering\arraybackslash}p{#1}}
\newcolumntype{M}[1]{>{\centering\arraybackslash}m{#1}}
\graphicspath{{images/}}
\usepackage{subfig}
\usepackage{epstopdf}
\usepackage{multirow}
\DeclareMathOperator*{\argmin } {arg\, min}

\usepackage{url}
\usepackage{comment}
\usepackage{enumitem}

\btasfinalcopy 

\ifbtasfinal\fi
\begin{document}

\title{Gender and Ethnicity Classification of Iris Images using Deep Class-Encoder}

\author{Maneet Singh$^1$, Shruti Nagpal$^1$, Mayank Vatsa$^{1,2}$, Richa Singh$^{1,2}$, Afzel Noore$^2$, and Angshul Majumdar$^1$\\
$^1$IIIT-Delhi, India, $ ^2$West Virginia University\\
\{\tt\small maneets, shrutin, mayank, rsingh, angshul\}@iiitd.ac.in, afzel.noore@mail.wvu.edu
}


\maketitle
\thispagestyle{empty}

\begin{abstract}
Soft biometric modalities have shown their utility in different applications including reducing the search space significantly. This leads to improved recognition performance, reduced computation time, and faster processing of test samples. Some common soft biometric modalities are ethnicity, gender, age, hair color, iris color, presence of facial hair or moles, and markers. This research focuses on performing ethnicity and gender classification on iris images. We present a novel supervised autoencoder based approach, Deep Class-Encoder, which uses class labels to learn discriminative representation for the given sample by mapping the learned feature vector to its label. The proposed model is evaluated on two datasets each for ethnicity and gender classification. The results obtained using the proposed Deep Class-Encoder demonstrate its effectiveness in comparison to existing approaches and state-of-the-art methods.
\end{abstract}

\section{Introduction}
\begin{figure}
\centering
\includegraphics[width = 2.5in]{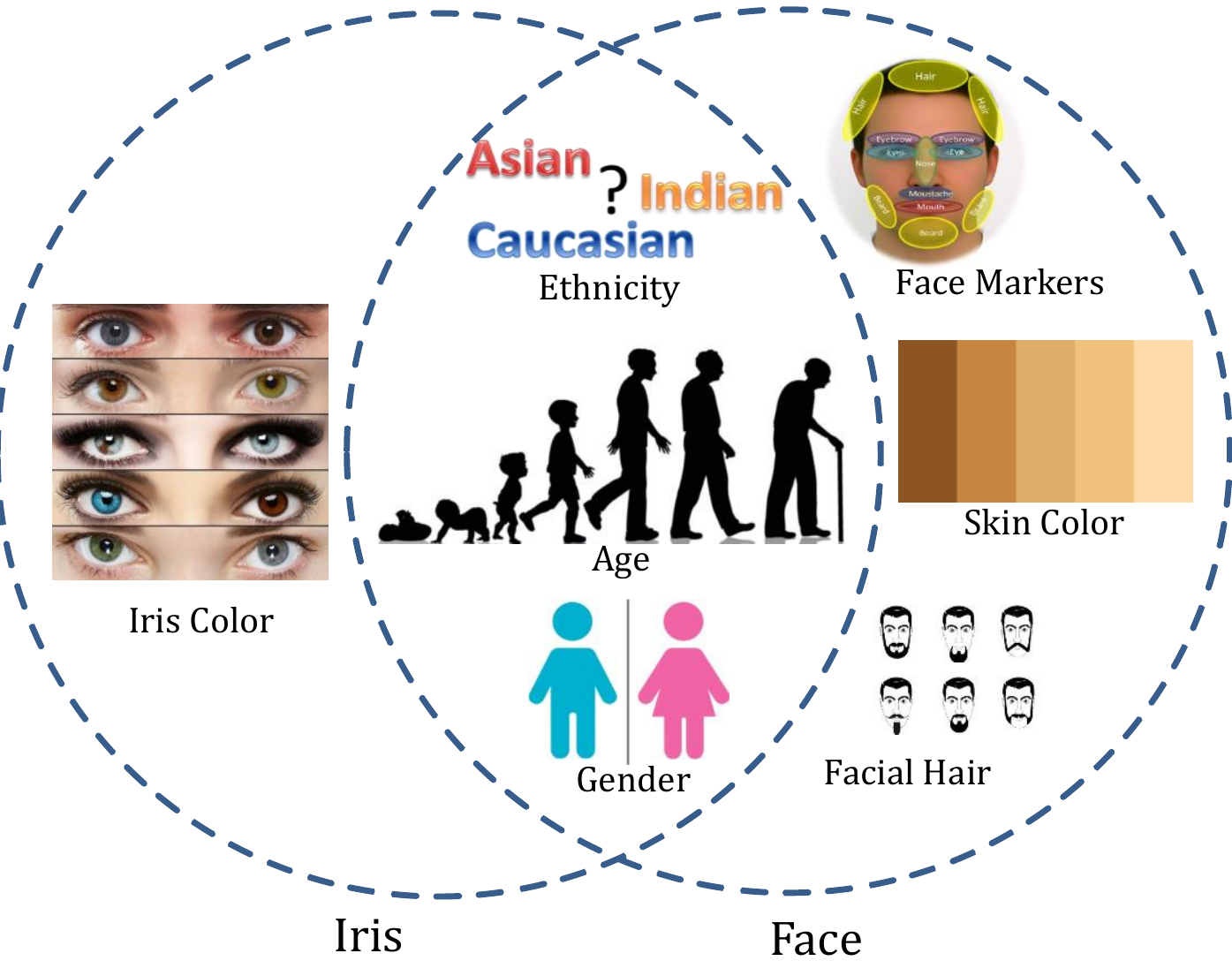}
\caption{Sample soft biometric modalities that can be extracted from face and iris images. All images have been taken from the Internet. }
\label{fig:intro}
\vspace{-10pt}
\end{figure}

Soft biometric traits are physical or behavioral characteristics that can be extracted from the human body which help differentiate individuals from one another. However, they often lack information that is permanent or sufficiently discriminative to uniquely identify an individual. These traits are valuable as they aid in recognition and enhance the performance of automated biometric systems by reducing the search space. This results in a significant decrease of testing time as well. Some of the most commonly used soft biometric modalities are gender, ethnicity, gait, age, height, hair color, eye color, and presence of moles or mustache on the face \cite{ross, review2}. Figure \ref{fig:intro} presents few sample soft biometric modalities that can be extracted from face and iris. 

Iris texture has emerged as one of the most robust and widely used biometric modality for automated identity recognition \cite{ishan}. Studies in literature have also established the presence of similar information in iris patterns to discriminate between ethnicities and gender \cite{ross, study, bow, tapia, NDgFI}. That is, samples belonging to the same ethnicity, have some commonalities in the iris pattern, and the ones belonging to the same gender also share certain information.

In literature, gender from iris images was first predicted by Thomas \textit{et al.} \cite{thomas} using an in-house dataset of left irises only. The iris images were segmented and normalized, onto which log-Gabor filter was applied to obtain the iris texture information. Features like mean, standard deviation, and local variation were computed for different neighborhoods of the texture pattern. Seven additional geometric features were also computed, followed by projection onto a random subspace ensemble of trees for classification. 
Lagree and Bowyer \cite{bow} extended this work by computing texture features for different horizontal and vertical bands of the normalized iris image. A total of 630 features were computed for 14 regions in a given image. These features were used as input to the Sequential Minimum Optimization algorithm for classification. The same extracted features were also used to perform ethnicity classification of iris images. Tapia \textit{et al.} \cite{tapia} used uniform LBP with concatenated histograms to perform gender classification on iris images on an in-house UND dataset. However, in this work, no mutual exclusivity of subjects was maintained between the training and testing partitions. Building upon this, in 2016 \cite{NDgFI}, the disjoint train-test sets in terms of the subject were created, and gender classification was performed using the same binary texture pattern that is used for identity recognition. This work presented results on three datasets, the UND dataset used in \cite{tapia}, a novel ND-Gender From Iris (ND-GFI) dataset, and a subject disjoint validation set (UND\_V).

Qiu \textit{et al.} \cite{qiu} first proposed the idea of classifying iris images into ethnicity based categories. The authors combined images from three datasets and performed classification between Asian and Non-Asian by computing the Gabor energy ratio between two regions of the iris. 
Continuing this work, a method to represent iris images using iris textons was developed \cite{textons}. The authors performed classification between Asian and Non-Asian on the CASIA BioSecure dataset. A filter bank of 40 Gabor filters was created and a 40-dimensional feature vector was computed. 
Each iris image was represented with a histogram of iris textons, which was used as a feature for SVMs to perform classification. The authors reported an accuracy of 91.02\% to classify Asians v/s non-Asians. However, it is essential to note that these results were demonstrated on images collected from different datasets for different ethnicities, thus incorporating inter-database variations at the time of classification. Lagree and Bowyer \cite{bow} proposed to identify the ethnicity using features obtained from several bands for 14 subregions of the iris. In 2012, Zarei and Mou \cite{last} used artificial neural networks to predict the ethnicity. The authors computed a total of 882 features from various regions of the iris, which were combined to create a feature vector used as input to a neural network for classification. 

This research presents ethnicity and gender classification from iris images, using the proposed \textit{Deep Class-Encoder}. It is an autoencoder based supervised model which utilizes the robust feature extraction capabilities of deep learning and discriminative capabilities of supervised models. The research contributions of this work are two-fold:
\begin{itemize}[leftmargin=*]
\item An autoencoder based supervised model, \textit{Deep Class-Encoder} is presented. The proposed model incorporates supervision by learning weights such that the hidden representation is mapped onto its class label. The proposed model is optimized using Alternating Direction Method of Multipliers \cite{opt}, instead of a gradient descent approach, thus reducing the training time substantially. 
\vspace{-5pt}
\item For gender classification, results are reported on existing protocols of ND-Iris-0405 \cite{ndIris} and ND-Gender From Iris \cite{NDgFI} datasets. Ethnicity classification is performed on ND-Iris-0405 and the proposed Multi-Ethnicity dataset. A new protocol is also defined on the Multi-Ethnicity dataset for ethnicity classification. The proposed model achieves state-of-the-art results for each task. 
\end{itemize}
\section{Proposed Deep Class-Encoder}
\label{prop}
The success of unsupervised feature learning architectures has been well established over the past couple of years, especially with the advent of representation learning techniques such as Deep Learning \cite{rep}. The performance of these architectures is not restricted to a particular domain, and spans across areas such as speech recognition, biometric recognition, computer vision, and natural language processing \cite{deep}. Unsupervised architectures aim to harness the information provided from unlabeled data to learn representative features used for classification at a later stage.

Autoencoders are deep learning architectures used to learn representations in an unsupervised manner. A traditional autoencoder consists of two sub-systems, encoder and decoder. For a given input $\mathbf{X}$, the encoder is used to obtain the representation $\mathbf{H}$ using the learned encoder weights $\mathbf{W_{e}}$. The decoder is used for reconstructing the input from the learned representation using the decoder weights, $\mathbf{W_{d}}$. The loss function of an autoencoder is optimized by minimizing the difference between the given input data, $\mathbf{X}$, and the reconstructed data, $\mathbf{\hat{X}}$. It is mathematically expressed as:
\begin{equation} \label{AE}
\min \left \|\mathbf{X} - \mathbf{\hat{X}} \right \|_{F}^{2}
\end{equation}
where, the input consists of $n$ training samples, the hidden representation is computed as $\mathbf{H} = \phi \mathbf{(W_{e}X)}$, and the reconstructed data $\mathbf{\hat{X}} = \mathbf{W_{d}H}$. $\phi$ is an activation function which can be any non-linear function such as $sigmoid$, or $tanh$ or linear activation corresponding to unit activation. Upon expanding Equation \ref{AE}, with unit activation function, the loss function can be expressed as:
\begin{equation} \label{ExpandedAE}
\mathbf{
\argmin_{\textit{$\mathbf{W_d,W_e}$}} \left \| \mathbf{X - W_{d}W_{e}X }\right \|_{F}^{2} }
\vspace{-5pt}
\end{equation}

A traditional biometrics pipeline consists of feature extraction, followed by classification. The feature extraction module extracts meaningful representations from the input, while the classifier learns a boundary in order to distinguish between the representations. Autoencoders have been used as unsupervised models for feature extraction in several applications. In order to learn meaningful representations for specific tasks, researchers have proposed incorporating supervision in the autoencoder model. Most of the existing supervised models do not explicitly encode the class label, but encode only the class information (same/different) for feature learning \cite{supervised, lcsse, contrastive}. Encoding class information often leads to reducing the intra-class variability.
 
\begin{figure}[]
\centering
  \subfloat[Traditional Unsupervised Autoencoder]{\includegraphics[width=2.7in,  trim={1.2cm, 0cm, 0.1cm, 0.0cm}]{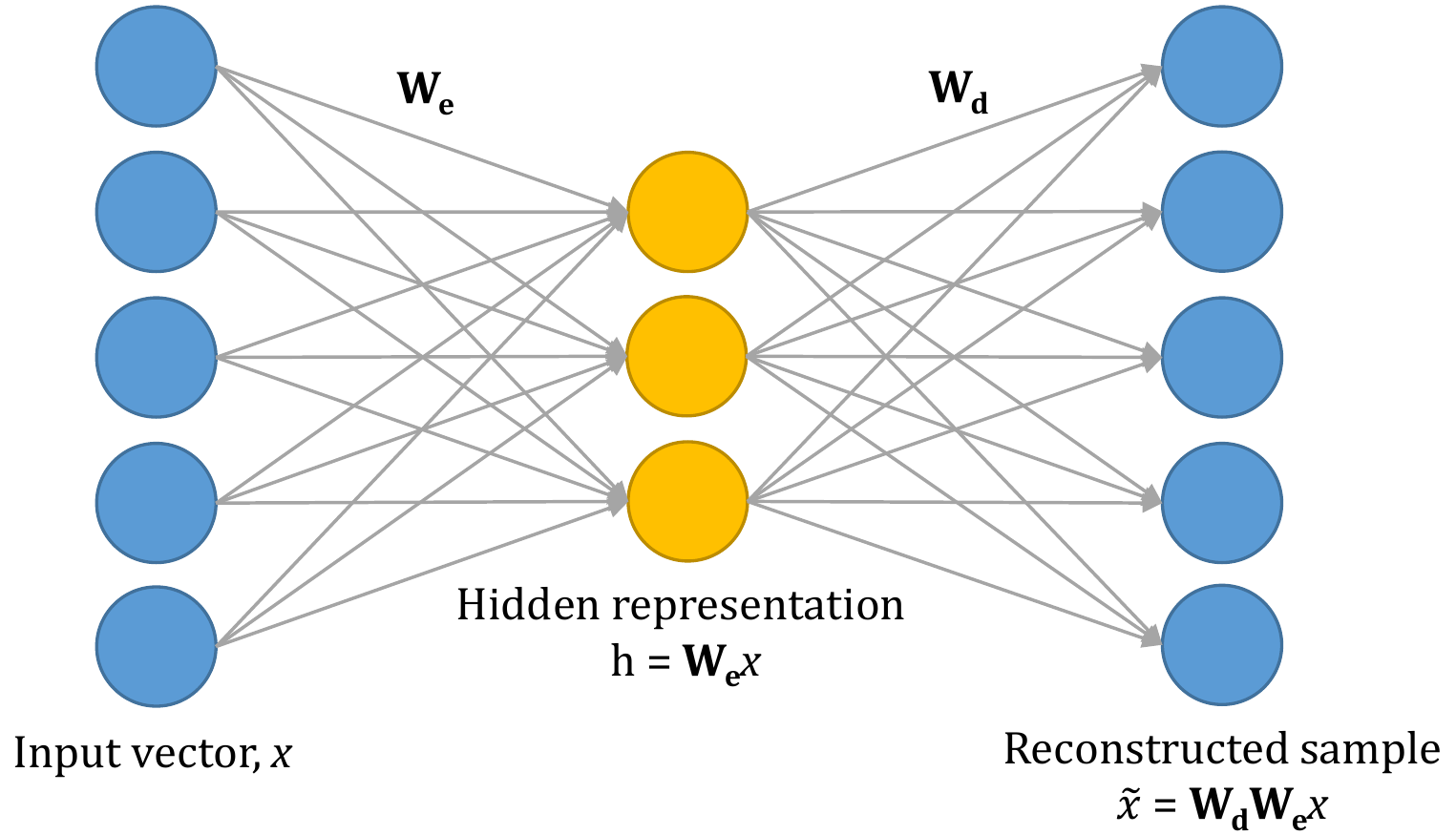}} \\
  \subfloat[Proposed Class-Encoder]{\includegraphics[width= 2.7in, trim={1.1cm, 0cm, 0.1cm, 0.5cm}]{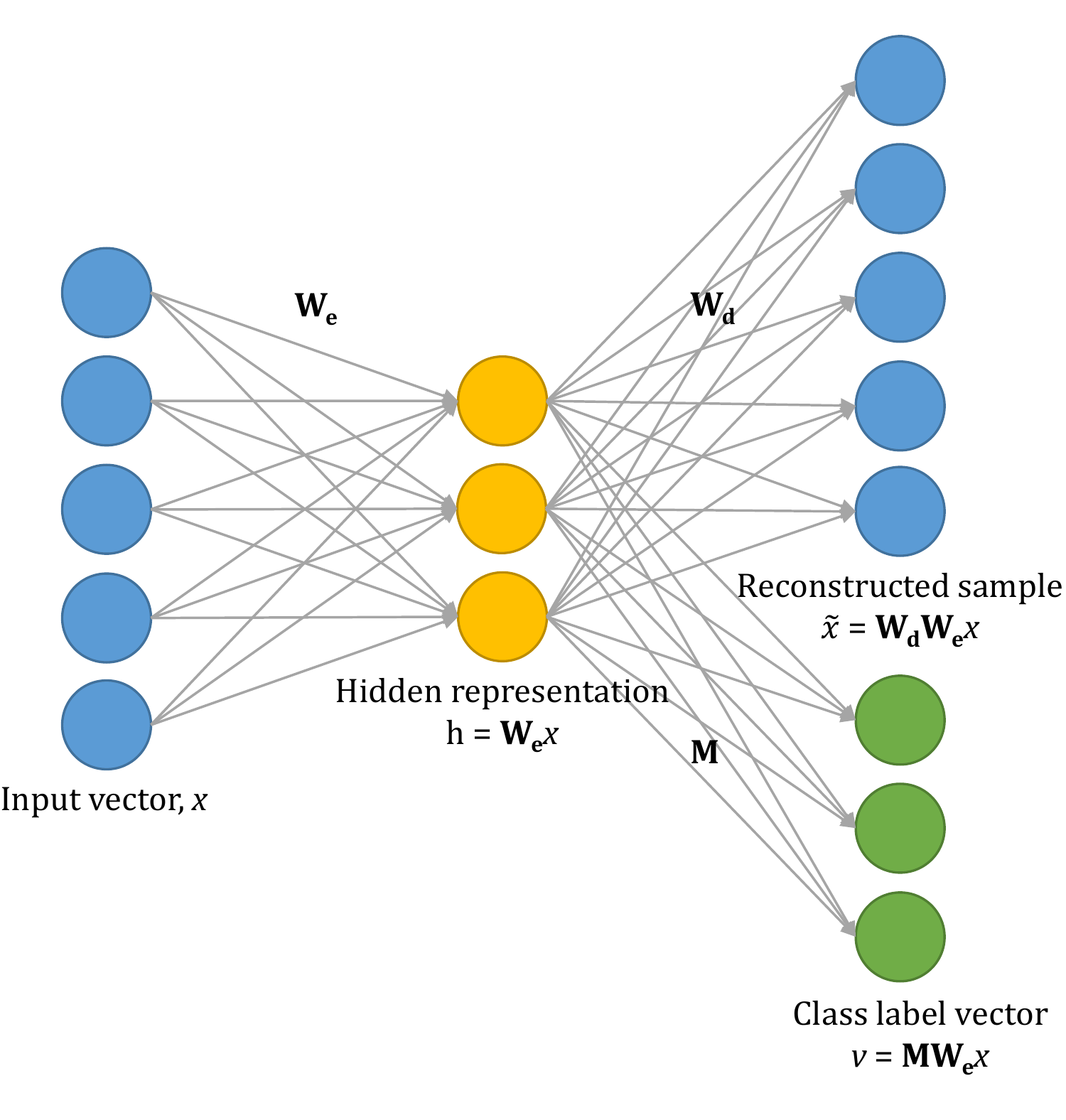}} \\
  \caption{$\mathbf{W_e}$ and $\mathbf{W_d}$ correspond to the encoding and decoding weights respectively, while $\mathbf{M}$ captures the linear mapping between the feature vector and its class label.}
\label{fig:algo}
\vspace{-10pt}
\end{figure}

In this research, we incorporate supervision in the feature extraction module in an attempt to introduce discriminability at the time of feature extraction itself. It is our hypothesis that feature extraction from such a model would facilitate better classification. We propose a model which utilizes the robust feature extraction capability of autoencoders, while learning discriminative features to enhance classification. The proposed architecture, termed as \textit{Class-Encoder} is built such that the class label of a given training sample is encoded in the learned feature representation, thereby making the model supervised in nature. This is done by incorporating an additional mapping matrix $\mathbf{M}$, which maps the hidden representation $\mathbf{W_eX}$ onto it's corresponding class label $\mathbf{C}$. Mathematically, it can be expressed as:
\begin{equation} \label{reg}
\mathbf{
\min_{\mathbf{M}} \left \| C - MW_{e}X \right \|_{F}^{2}}
\vspace{-5pt} 
\end{equation}
where, $\mathbf{C} \in R^{ l\times n}$ ($l$ being the number of distinct classes) is a binary vector with its $i^{th}$ element set to 1 if the data sample belongs to class $i$, rest being zero. Matrix $\mathbf{M}$ captures the linear mapping between the feature vector ($\mathbf{W_{e}X}$) and it's corresponding class label. Thus, under a fixed $\mathbf{M}$ and $\mathbf{W_e}$, samples belonging to the same class should map to the same class label. Extending Equation \ref{ExpandedAE} to incorporate the above term, the loss function of the proposed Class-Encoder can thus be given as:
\begin{equation} \label{lcaeMatrix} \mathbf{
\min_{\textit{$\mathbf{M,W_d,W_e}$}} \left \|X - W_{d}W_{e}X \right \|_{F}^{2} + \lambda \left \| C - MW_{e}X \right \|_{F}^{2} }
\vspace{-5pt}
\end{equation}
where, the regularization parameter, $\lambda$, controls the relative contribution of the two terms. In Equation \ref{lcaeMatrix}, the first term corresponds to the loss function of the traditional autoencoder. It facilitates learning of parameters such that the reconstruction error is minimized and a meaningful representation is learned. The second term incorporates supervision in the autoencoder formulation and enables the model to learn a discriminative representation of the input data. The weight matrix is learned such that upon projecting the input data on it, the representation maps to a specific class label, thus encoding class-specific information in the feature learning process. Figure \ref{fig:algo} gives an overview of the proposed model, and it's comparison with the traditional unsupervised autoencoder.

Equation \ref{lcaeMatrix} represents a least square formulation, however, the large size of matrices involved and the non-convexity of the problem increases the computational costs. To mitigate the same, the Majorization-Minimization (MM) technique \cite{MM} along with Alternating Direction Method of Multipliers (ADMM) \cite{opt} is utilized. The aim of the MM technique is to replace complex equations with simpler and easier optimization steps. Moreover, ADMM does not involve computing the derivatives at each epoch, which leads to significant lower training time as compared to a traditional Stacked Autoencoder. 

For a $k$ layer model of Deep Class-Encoder, Equation \ref{lcaeMatrix} can be extended as follows:
\begin{equation} \label{lcaeMatrixDeep} 
\begin{gathered} \mathbf{
\min_{\textit{$\mathbf{M,W_d,W_e}$}} \left \|X - (W_{d}^1W_{d}^2...W_{d}^{\textit{k}}(W_{e}^{\textit{k}}...W_{e}^2W_{e}^1X)) \right \|_{F}^{2}} + \\
\sum_{i=1}^k\mathbf{\lambda \left \| C - M^{\textit{i}}(W_{e}^{\textit{i}}X^{\textit{i}}) \right \|_{F}^{2}}
\end{gathered}
\end{equation}
where, $\mathbf{W_e^{\textit{k}}}$, $\mathbf{W_d^{\textit{k}}}$, and $\mathbf{M^{\textit{k}}}$ correspond to the encoding weights, decoding weights and the mapping matrix of the $k^{th}$ layer, respectively. $\mathbf{X^{\textit{k}}}$ refers to the input to the $k^{th}$ layer of the model, which is defined as $\mathbf{W_{e}^\textit{{k-1}}X^{\textit{{k-1}}}}$. At the first layer, $\mathbf{X^1} = \mathbf{X}$, that is, the input data. Thus, for a $k$-layer model, the proposed Class-Encoder aims to learn discriminative representations at each layer by modeling the class label information during the feature learning process.

\section{Ethnicity and Gender Classification using Deep Class-Encoder}
\label{exp}
\begin{figure*}
\centering
\subfloat[ND-GFI Dataset]{\includegraphics[trim={0cm, 0cm, 0cm, 1cm}, width = 5cm]{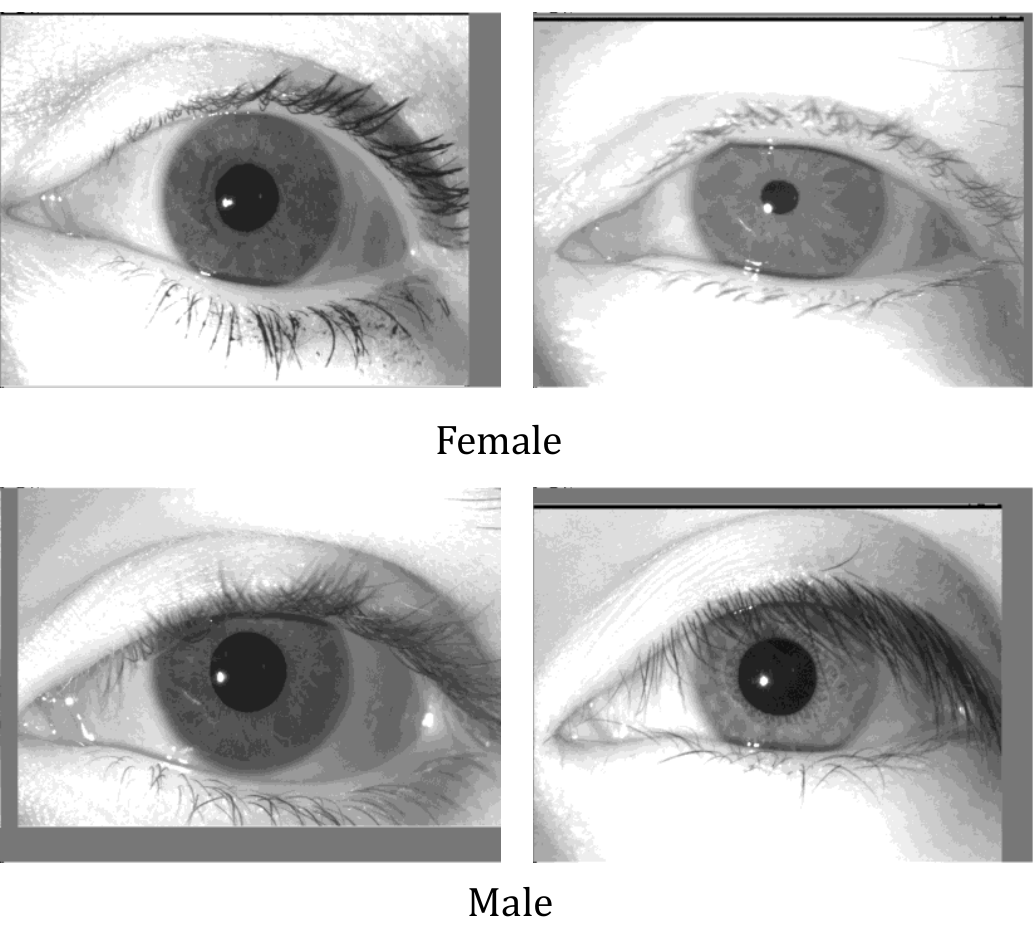}} 
\subfloat[ND-Iris-0405 Dataset]{\includegraphics[trim={0cm, 0cm, 0cm, 1cm}, width = 5cm]{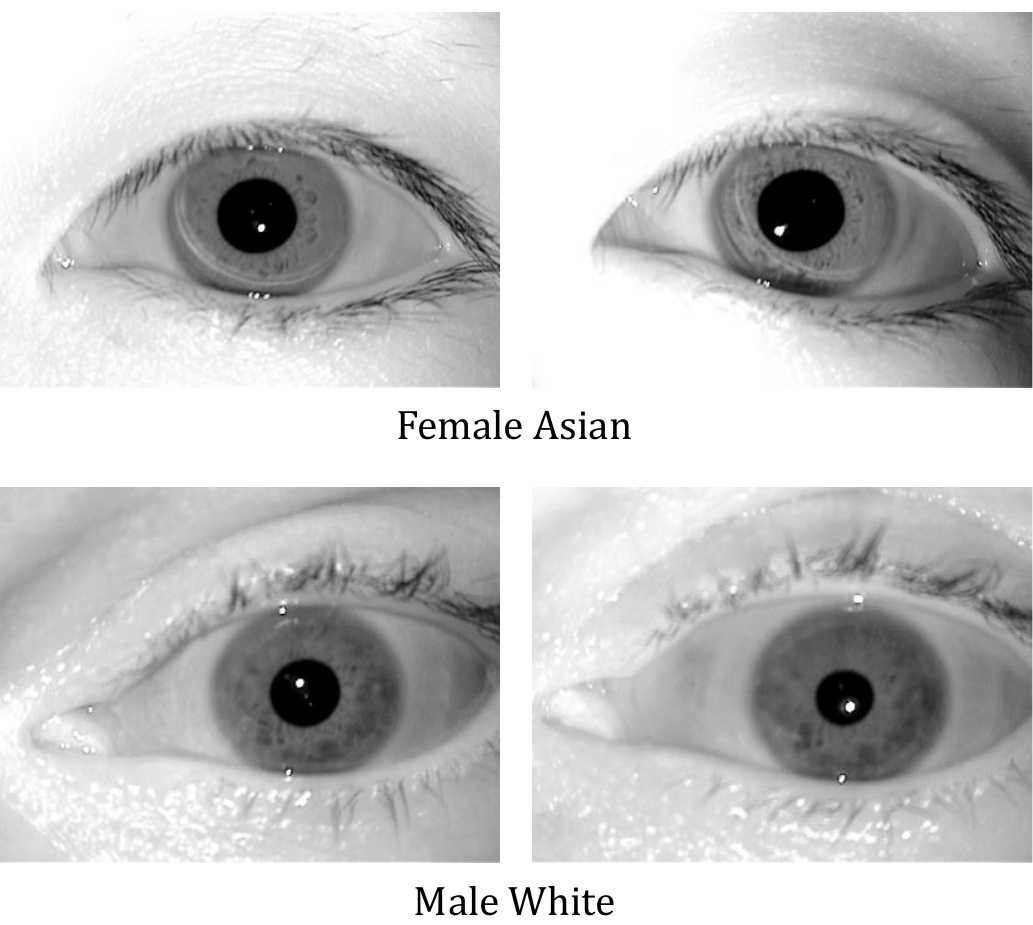}} 
\subfloat[Multi-Ethnicity Iris Dataset]{\includegraphics[trim={0cm, 0cm, 0cm, 1cm}, width = 5cm]{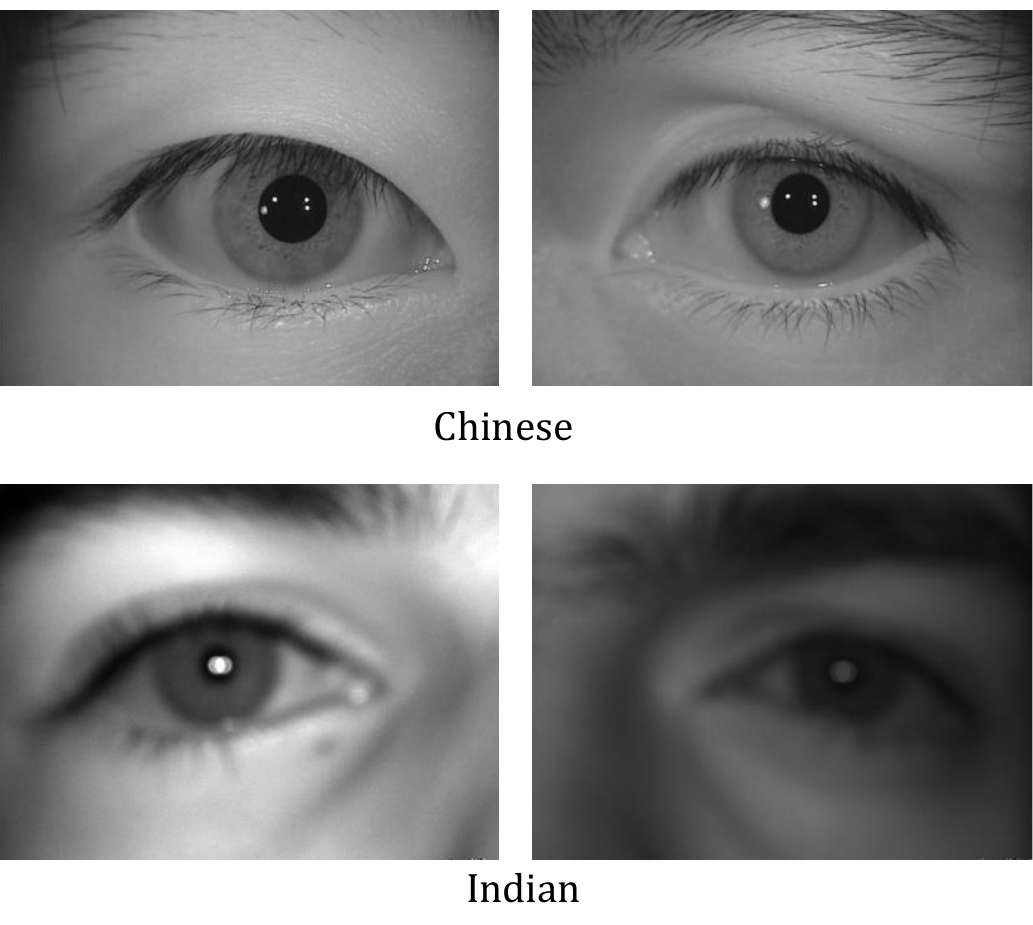}} 
\vspace{-5pt}
\caption{Sample images from the three datasets used for performing gender and ethnicity classification on iris images.}
\label{fig:sample}
\end{figure*}

Deep Class-Encoder is used for the task of gender and ethnicity classification of iris images. For both the problems, experiments are performed on two datasets each. Details regarding the algorithm, datasets used, and experimental protocols are given below.
\subsection{Ethnicity Classification}
\label{sec:eth}
The proposed \textit{Deep Class-Encoder} is used to perform ethnicity classification using Equation \ref{lcaeMatrixDeep}. Here, $\mathbf{X}$ is a $m\times n$ matrix which contains the input images, where $m$ is the size of the input image and $n$ is the number of input samples. In this case, iris images captured in the NIR spectrum are given as input, resized to $48 \times 64$, therefore $m$ being equal to 3,072. $\mathbf{W_d, W_e}$ are the weight parameters learned at the time of training. $\mathbf{C}$ is the class label matrix of dimension $l \times n$, where $l$ is the number of distinct classes. 
A two layer Deep Class-Encoder of dimension [$3072$, $768$, $768$] is learned. In order to evaluate the efficacy of the proposed model, results and comparison have been shown with two classifiers: Random Decision Forest (RDF) and a two layer Neural Network (NNet) having dimension  [$768$, $192$, $48$].

\subsubsection{Datasets and Experimental Protocol}
Two datasets, ND-Iris-0405 \cite{ndIris} and a combined Multi-Ethnicity Iris Dataset, have been used in this research. \\
\textbf{ND-Iris-0405 dataset} consists of 64,980 images corresponding to 356 subjects, out of which 158 subjects are Caucasian (White), 82 are Asian, and the remaining 24 belong to other ethnicities. Since the number of samples belonging to \textit{other} ethnicity are very limited, images corresponding to the Caucasian and Asian subjects have been chosen for evaluation, thereby resulting in a two-class problem (Caucasian or Asian), consisting of 60,259 images. Out of these, 26,272 images (containing equal number of Caucasian and Asian images) are used for training, while the remaining images form the test set.

\noindent\textbf{Multi-Ethnicity Iris dataset} is created by combining three existing datasets, due to the lack of publicly available iris datasets containing images of multiple ethnicities. The dataset consists of images from:
\begin{itemize} [leftmargin=*]
\item CASIA-Iris V3 dataset \cite{casiaV3} collected by the Chinese Academy of Sciences' Institute of Automation (CASIA): The dataset consists of three parts, CASIA-Iris-Interval, CASIA-Iris-Lamp, and CASIA-Iris-Twin. Out of these, CASIA-Iris Lamp and CASIA-Iris Twin contain 18,197 images of only Chinese ethnicity, that have been used. 
\vspace{-5pt}
\item IMP dataset \cite{imp}: This dataset contains images collected in the visible and NIR spectrum, and night-time mode. Out of these, 630 images of Indian ethnicity captured in the NIR spectrum are used.
\vspace{-5pt}
\item ND-Iris-0405 dataset \cite{ndIris}: As mentioned previously, this dataset contains images of subjects belonging to Asian, Caucasian, or other ethnicities. Since Asian might correspond to Chinese or Indian as well, and further categorization has not been provided in the dataset, we only use 41,518 images pertaining to Caucasian subjects in the Multi-Ethnicity Iris dataset.   
\end{itemize}
Therefore, the Multi-Ethnicity Iris dataset consists of 60,310 images of Chinese, Indian, and Caucasian ethnicities. 1,302 images are chosen for creating the training partition, which consists of equal number of images from all three classes (Chinese, Indian, and Caucasian), and the remaining images are used to create the test set.

\begin{table}[]
\centering
\caption{Experimental protocols for ethnicity and gender classification.}
\label{tab:exp}
\vspace{-5pt}
\begin{tabular}{|m{2.8cm}|P{2.3cm}|P{2.2cm}|}
\hline
\textbf{Dataset} & \textbf{No. of Training Images} & \textbf{No. of Testing Images} \\
\hline
\hline
\multicolumn{3}{|c|}{Ethnicity Classification} \\
\hline
ND-Iris-0405 \cite{ndIris} & 26,272 & 33,987 \\
\hline
Multi-Ethnicity Iris & 1,302 & 59,008 \\
\hline
\multicolumn{3}{|c|}{Gender Classification} \\
\hline
ND-Iris-0405 \cite{ndIris} & 42,899 & 22,081 \\
\hline
ND-GFI \cite{NDgFI} & 2,399 & 600 \\
\hline
UND\_V \cite{NDgFI} & - & 1,944 \\
\hline
\end{tabular}
\vspace{-10pt}
\end{table}

\subsection{Gender Classification}
Deep Class-Encoder is also used for performing gender classification on iris images. Similar to the previous experiments, the images are down-sampled to $48\times64$. Thus, $\mathbf{X}$ is of dimension $3072\times n$, where $n$ is the number of input images. $\mathbf{C}$ is a $2\times n$ binary-matrix containing the class labels. The architecture of the feature extractors and classifiers is kept consistent with that described in Section \ref{sec:eth}.
\subsubsection{Datasets and Experimental Protocol}
Experiments have been performed on two datasets: ND-Iris-0405 \cite{ndIris} and ND-Gender From Iris (GFI) \cite{NDgFI}. \\
\textbf{ND-Iris-0405 dataset} consists of 64,980 images corresponding to 158 females and 198 males. 70\% data of each class is used for training (resulting in 42,899 images), while the remaining form the test set. \\
\textbf{ND-Gender From Iris (ND-GFI) dataset} consists of 3,000 images from 750 males and 750 females. The dataset consists of a pre-defined protocol \cite{NDgFI}, where 80\% of the data (per class) is used for training, and the remaining 20\% is used for testing. A separate subject-disjoint set (UND\_V) containing 1,944 images has also been provided for evaluating the performance of the trained model. The same protocol is followed for the evaluation of the proposed model. All protocols ensure mutually exclusive training and testing sets, such that there is no image which occurs in both the partitions. Figure \ref{fig:sample} presents sample images from all three datasets, and Table \ref{tab:exp} summarizes the protocols followed for the experimental evaluation. 

\begin{table}[]
\centering
\caption{Ethnicity classification accuracy (\%) on the ND-Iris-0405 dataset with two classifiers: Random Decision Forest (RDF) and Neural Network (NNet).}
\vspace{-5pt}
\begin{tabular}{|l|P{1cm}|P{1cm}|}
\hline
\textbf{Algorithm} & \textbf{RDF} & \textbf{NNet} \\
\hline
\hline
Stacked Autoencoder \cite{sae} & 87.24  & 80.05 \\
\hline
Stacked Denoising Autoencoder \cite{sdae} & 85.20 & 64.51 \\
\hline
Deep Belief Network \cite{dbn} & 85.12 & 87.43 \\
\hline
Discriminative RBM \cite{drbm} & \multicolumn{2}{c|}{90.33}  \\
\hline
\textbf{Proposed Deep Class-Encoder} & \textbf{89.35} & \textbf{94.33}\\
\hline
\end{tabular}
\label{tab:ndEthn}
\vspace{-10pt}
\end{table}

\section{Experimental Results and Analysis}
\label{res}
The proposed model has been compared with existing Deep Learning models for performing gender and ethnicity classification on iris images. Comparison has been performed with Stacked Autoencoder (SAE), Stacked Denoising Autoencoder (SDAE) \cite{sdae}, Deep Belief Network (DBN) \cite{dbn}, Discriminative Restricted Boltzmann Machine (DRBM) \cite{drbm}, and AlexNet \cite{alexnet}, i.e. a Convolutional Neural Network based model. To be consistent with the proposed model, feature extraction models (SAE, SDAE, DBN) have the same architecture as that of the proposed Deep Class-Encoder (described in Section \ref{sec:eth}). This is followed by learning a classifier (RDF or NNet) for classification. 
Owing to the class imbalance in the test samples, mean class-wise accuracy has been reported throughout this paper. Since no existing Commercial-Off-The-Shelf system predicts the attributes of gender and ethnicity for iris images, a comparison could not have been drawn for the same. Task specific analysis for the proposed Deep Class-Encoder model is given below.
 
\begin{table}[]
\centering
\caption{Ethnicity classification accuracy (\%) on the Multi-Ethnicity Iris dataset with two classifiers: RDF and NNet.}
\label{tab:multiEthn}
\vspace{-5pt}
\begin{tabular}{|l|P{1cm}|P{1cm}|}
\hline
\textbf{Algorithm} & \textbf{RDF} & \textbf{NNet} \\
\hline
\hline
Stacked Autoencoder \cite{sae} & 95.76  & 90.42 \\
\hline
Stacked Denoising Autoencoder \cite{sdae} & 95.13 & 87.3 \\
\hline
Deep Belief Network \cite{dbn} & 96.97 & 95.22 \\
\hline
Discriminative RBM \cite{drbm} & \multicolumn{2}{c|}{95.92}  \\
\hline
\textbf{Proposed Deep Class-Encoder} & \textbf{97.22} & \textbf{97.38}\\
\hline
\end{tabular}
\vspace{-10pt}
\end{table}

\subsection{Ethnicity Classification}
\begin{figure}
\centering
\includegraphics[ trim={0cm, 1cm, 0cm, 1cm}, width = 3.3in]{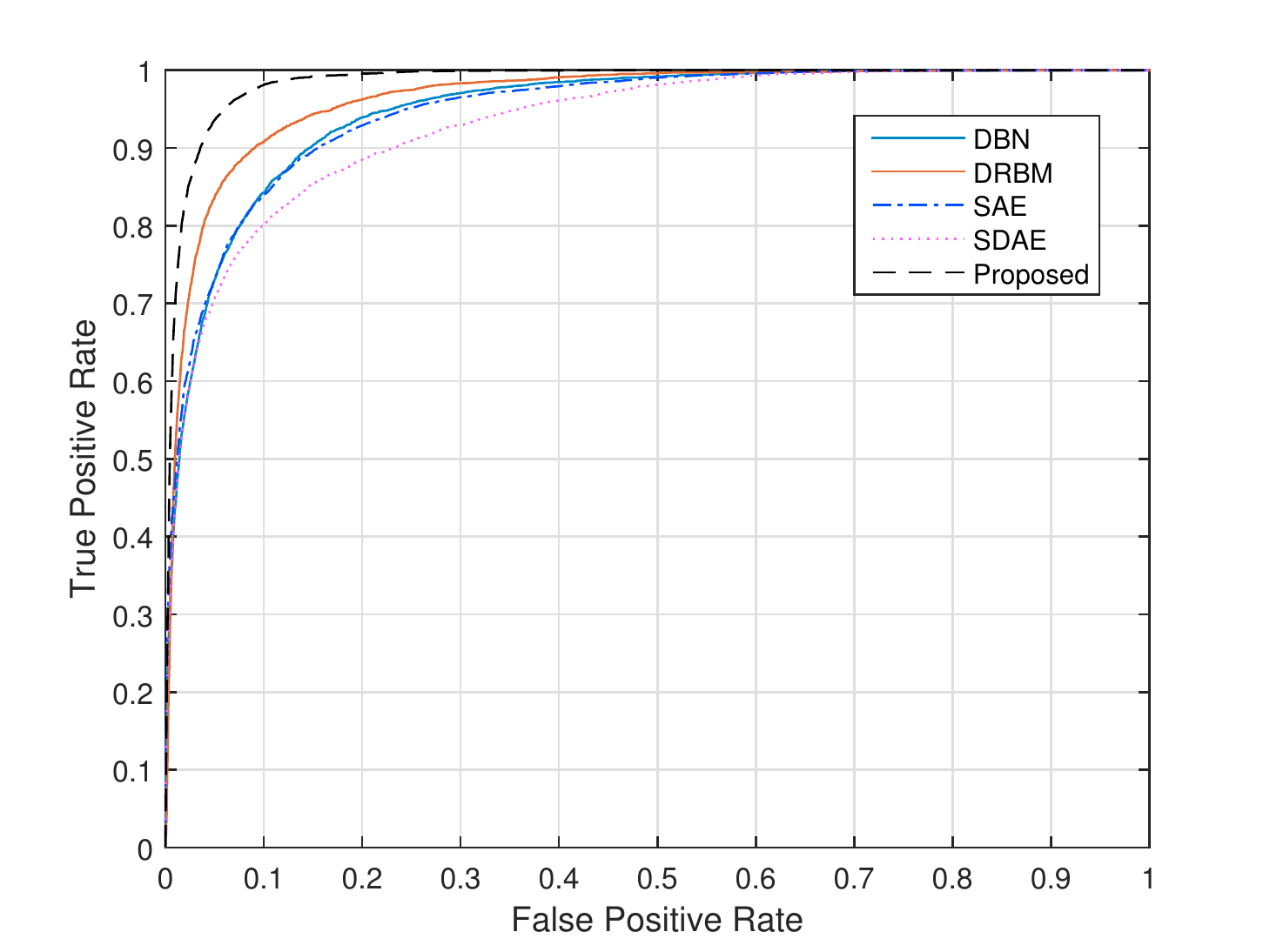}
\vspace{-5pt}
\caption{Ethnicity classification on the ND-Iris-0405 dataset. For all models, the best performance between RDF and NNet is plotted.}
\label{fig:ethRoc}
\vspace{-5pt}
\end{figure}

Tables \ref{tab:ndEthn} and \ref{tab:multiEthn} present the accuracy for ethnicity classification on ND-Iris-0405 and Multi-Ethnicity Iris dataset, respectively. Figure \ref{fig:ethRoc} also presents the Receiver Operative Characteristic (ROC) curve obtained on the ND-Iris-0405 dataset for ethnicity classification of Asians versus Caucasians. On the ND-Iris-0405 dataset, the proposed model achieves a classification accuracy of 89.35\% and 94.35\% with RDF and NNet classifier, respectively (Table \ref{tab:ndEthn}). Compared with other unsupervised feature learning models (SAE, SDAE, and DBM), Deep Class-Encoder presents an improvement of at least 7\%. Direct comparison with SAE (difference of around 14\% in accuracy) shows the benefit of encoding class labels in the proposed model.  Comparison with Discriminative RBM, which is a supervised Deep Learning architecture, also presents an improvement of 4\%. The key difference between Discriminative RBM and proposed Deep Class-Encoder is that the former models the joint probability of the sample and the label, while the later tries to learn representations such that under a fixed learned mapping, the hidden representations map to the class labels. It can thus be observed that the introduction of the mapping matrix facilitates learning of discriminative features. Table \ref{tab:conf} gives the confusion matrix obtained with the proposed Deep Class-Encoder and Neural Network. The low mis-classification rates for both the classes suggest that the proposed model is able to encode meaningful and discriminative representations across classes.

\begin{table}
\begin{center}
\caption{Confusion matrix of Deep Class-Encoder on the ND-Iris-0405 dataset for ethnicity classification.}
\label{tab:conf}
\vspace{-5pt}
\begin{tabular}{ | M{0.3cm} || M{1.6cm} | M{1.6cm}|M{1.6cm}|  } 
\hline
& \multicolumn{3}{c|}{\textbf{Predicted}} \\ 
\hline
\hline
\multirow{3}{*}{\rotatebox[origin=c]{90}{\textbf{Actual}}}&  & Asian & Caucasian \\
\cline{2-4}
& Asian & 93.78\% & 6.22\% \\
\cline{2-4}
& Caucasian & 5.12\% & 94.88\% \\
\hline
\end{tabular}
\end{center}
\vspace{-20pt}
\end{table}

\begin{figure}[t]
\centering
\includegraphics[width = 3.2in]{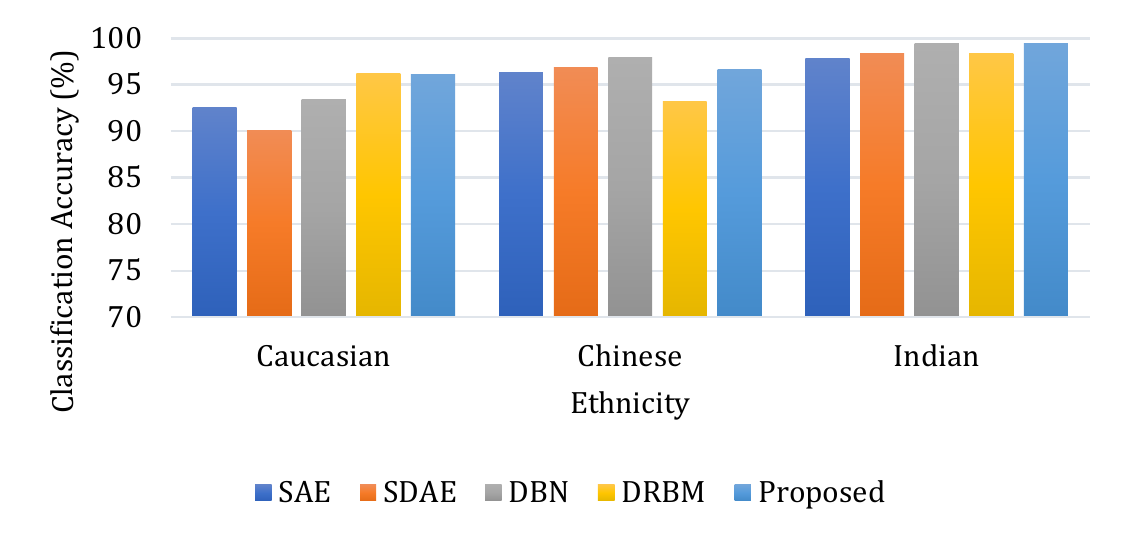}
\vspace{-5pt}
\caption{Classification performance (\%) of all algorithms on all three classes for the Multi-Ethnicity Iris dataset.}
\label{fig:ethBar}
\vspace{-5pt}
\end{figure}

\begin{figure}
\centering
\includegraphics[width = 3.2in]{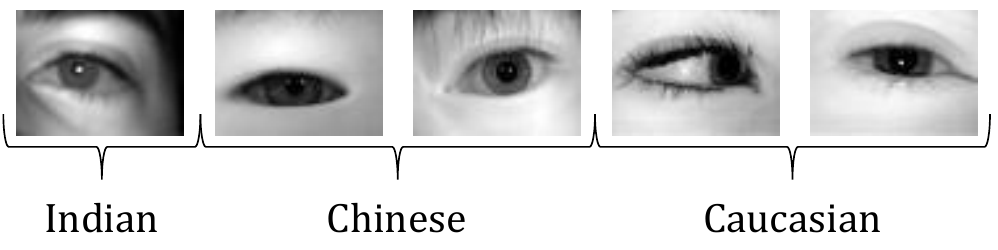}
\vspace{-5pt}
\caption{Sample images mis-classified by the proposed Deep Class-Encoder and Neural Network model for the Multi-Ethnicity Iris dataset. Artifacts such as hair (bangs) and incomplete iris information result in mis-classification.}
\label{fig:ethMis}
\vspace{-10pt}
\end{figure}

Similar results can be observed from Table \ref{tab:multiEthn} for ethnicity classification on Multi-Ethnicity Iris dataset. In this case, the classifier aims to classify the images as either Caucasian, Chinese, or Indian. The proposed model achieves a classification accuracy of 97.38\% (with Neural Network), which shows improvement over other comparative feature extraction models. Moreover, the variation in classification performance obtained with RDF and NNet for Deep Class-Encoder in within 1\%, whereas the difference is as high as 8\% for other comparative models (SDAE). This observation further instantiates that the proposed model learns robust features which are independent of the classifier used in the pipeline. Figure \ref{fig:ethBar} gives the bar graph with classification accuracies of all models for all three classes. The proposed model achieves at least 96\% classification accuracy for all three classes, thus promoting the feature learning process of Deep Class-Encoder. Interestingly, the model mis-classifies only \textit{one} sample of Indian ethnicity. As shown in Figure \ref{fig:ethMis}, overall illumination and size variations along with the presence of artifacts such as hair bangs render samples challenging for ethnicity classification. 

\begin{figure*}
\centering
\subfloat[ND-Iris-0405 Dataset]{\includegraphics[trim={0cm, 0cm, 0cm, 1.5cm}, width = 3.3in]{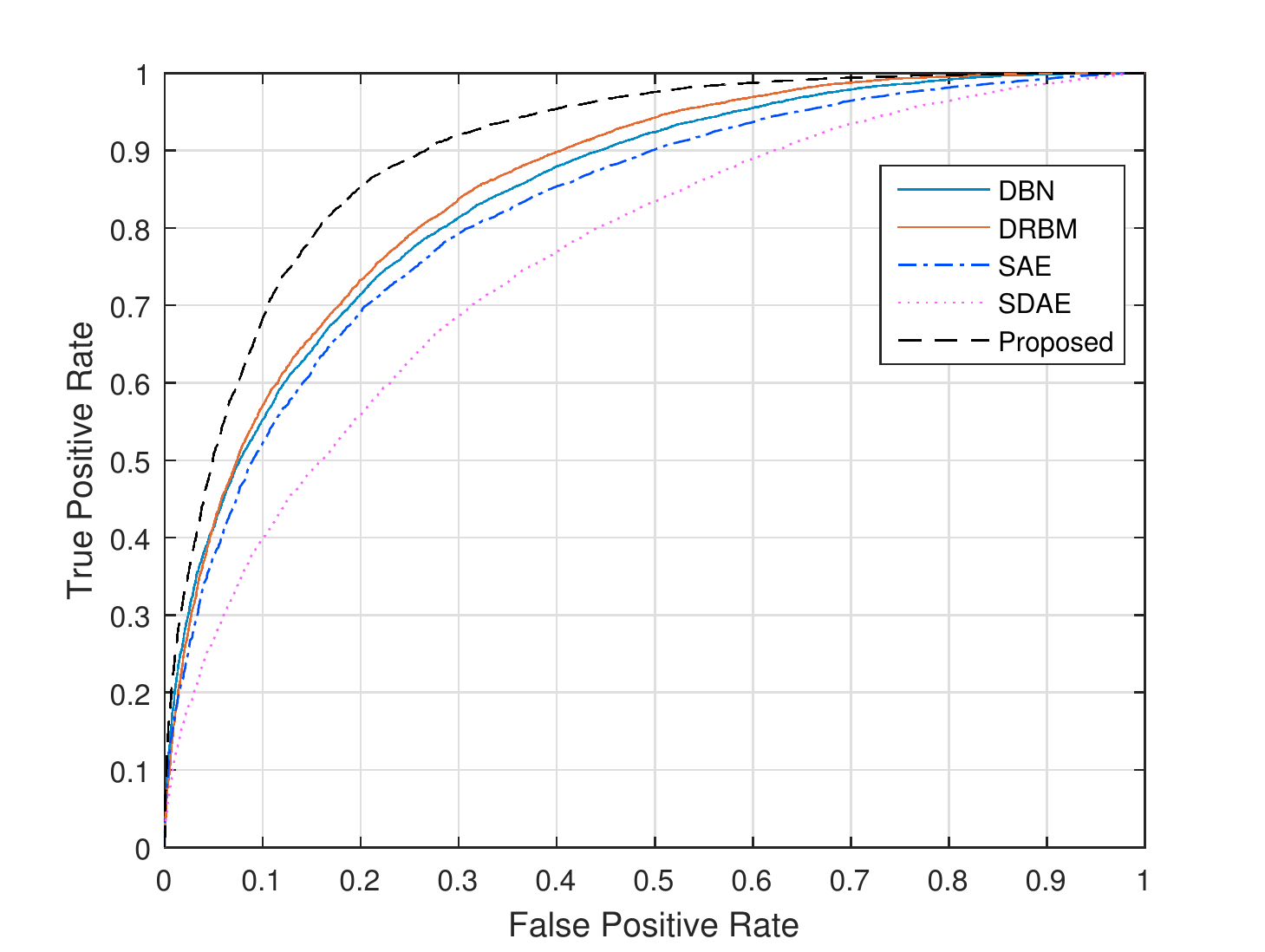}} 
\subfloat[ND-GFI Dataset]{\includegraphics[trim={0cm, 0cm, 0cm, 1.5cm}, width = 3.3in]{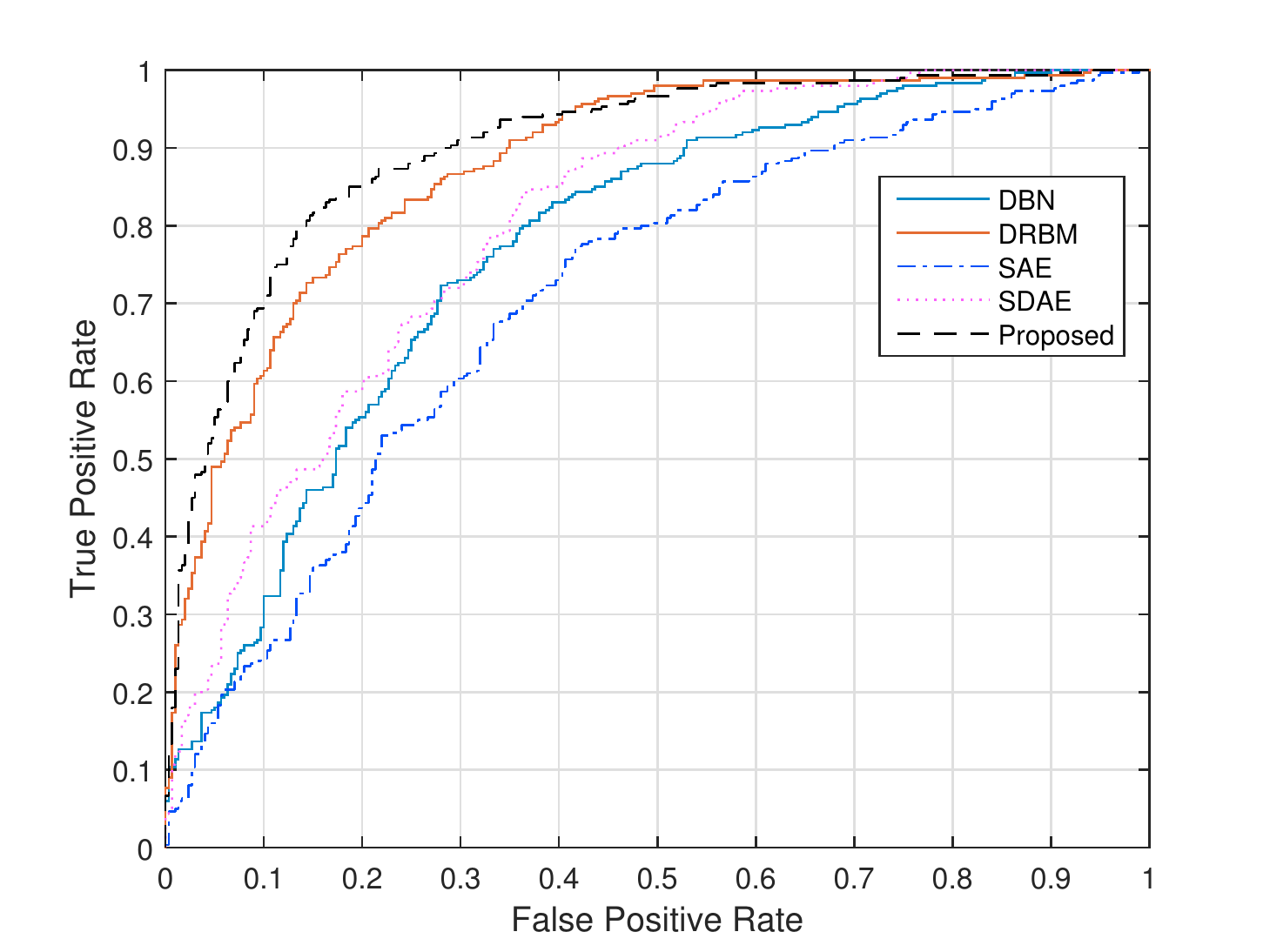}} 
\vspace{-5pt}
\caption{ROC curves for gender classification. For each model, the best performing classifier (RDF or NNet) is plotted.}
\label{fig:genderRoc}
\vspace{-10pt}
\end{figure*}

It is important to note that no direct comparison could be drawn with existing work on ethnicity classification, since there is no fixed protocol or datasets which are used for evaluation. While there exists works utilizing the ND-Iris-0405 dataset, however, experiments are performed on a very small subset of the dataset \cite{study}.

\subsection{Gender Classification}
\begin{figure}
\centering
\includegraphics[width = 3in]{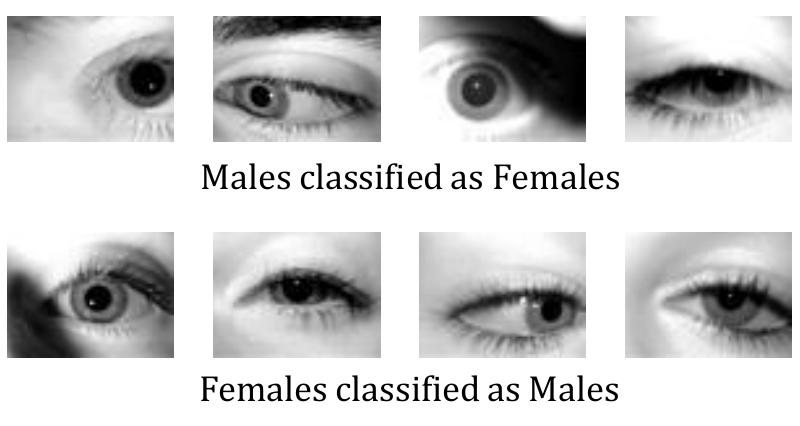}
\vspace{-10pt}
\caption{Sample mis-classifications by the proposed model. It can be observed that incomplete, poorly illuminated, and hidden iris tend to be more difficult to classify.}
\vspace{-10pt}
\label{fig:genMis}
\end{figure}

Tables \ref{tab:ndIris} and \ref{tab:ndGfi} present the gender classification accuracy on ND-Iris-0405, and ND-GFI dataset, respectively. Figure \ref{fig:genderRoc} also presents the ROC curves obtained on both the datasets. On the ND-Iris-0405 dataset, the proposed Deep Class-Encoder achieves a classification accuracy of 82.53\% (with Neural Network), which is at least 5.5\% better than other comparative models. Figure \ref{fig:genMis} presents sample images mis-classified by the proposed model. It can be observed that iris images having poor illumination or partial information serve as challenging samples. 

Experiments are also performed on the ND-GFI dataset, where, the Deep Class-Encoder achieved an accuracy of 83.17\%, showing at least 5\% improvement over other deep learning based comparative models. In literature, Tapai \textit{et al.} \cite{NDgFI} achieve an accuracy of 84.83\% on the same protocol, which means that the proposed model performs 10 extra mis-classifications. The proposed Deep Class-Encoder achieves a classification accuracy of 79.25\% on the UND\_V dataset. This dataset is provided with the ND-GFI dataset and models the real world scenario of disjoint subjects in the training and testing set. The protocol mentioned by the authors \cite{NDgFI} has been followed for evaluation, using which they report a classification accuracy of 77.5\% (state-of-the-art for this dataset). This results in a difference of 34 correctly classified samples, thereby promoting the generalization abilities of Deep Class-Encoder. 

\begin{table}[]
\centering
\caption{Gender classification accuracy (\%) on the ND-Iris 0405 dataset with two classifiers: RDF and NNet.}
\label{tab:ndIris}
\vspace{-5pt}
\begin{tabular}{|l|P{1cm}|P{1cm}|}
\hline
\textbf{Algorithm} & \textbf{RDF} & \textbf{NNet} \\
\hline
\hline
Stacked Autoencoder \cite{sae} & 74.73  & 66.71 \\
\hline
Stacked Denoising Autoencoder \cite{sdae} & 69.23 & 56.07 \\
\hline
Deep Belief Network \cite{dbn} & 75.74 & 76.03 \\
\hline
Discriminative RBM \cite{drbm} & \multicolumn{2}{c|}{70.76}  \\
\hline
\textbf{Proposed Deep Class-Encoder} & \textbf{80.06} & \textbf{82.53}\\
\hline
\end{tabular}
\vspace{-5pt}
\end{table}

\begin{figure} []
\centering
\includegraphics[width = 3in]{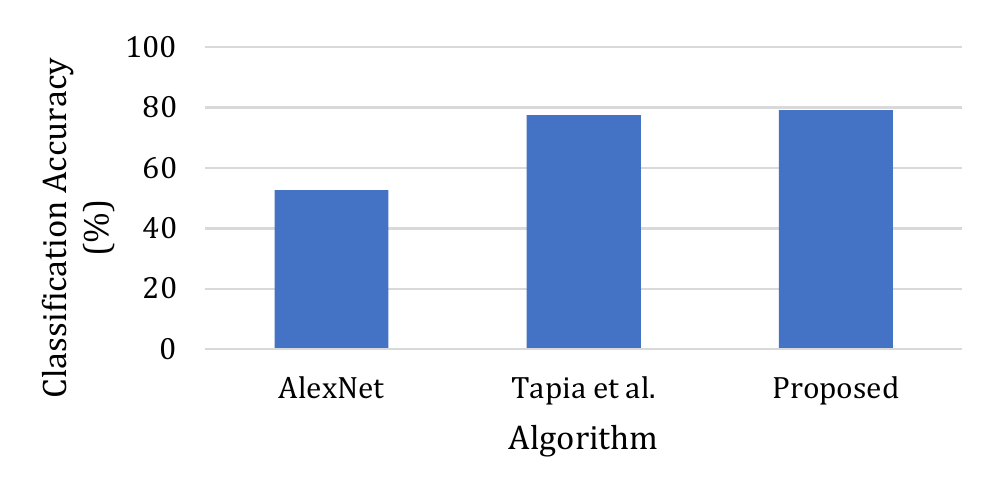}
\vspace{-10pt}
\caption{Bar graph presenting the gender classification accuracy obtained on the UND\_V dataset \cite{NDgFI}. }
\label{fig:Und}
\vspace{-10pt}
\end{figure}

Figure \ref{fig:Und} presents a bar graph with the classification accuracies of a CNN based feature extractor, AlexNet \cite{alexnet}, Tapia et al. \cite{NDgFI}, and the proposed model. Comparison with Convolutional Neural Network based feature extractor, AlexNet \cite{alexnet} on the UND\_V dataset further promotes the utility of the proposed model. Upon using the pre-trained model for feature extraction only, the proposed model achieves at least 16\% improved performance. A major advantage of the proposed Deep Class-Encoder is in terms of the training time. As compared to Stacked Autoencoders, Deep Class-Encoder takes one fourth the total training time. This computational advantage is primarily due to the Majorization-Minimization and Alternating Direction Method of Multipliers based optimization which does not involve computing derivatives at each epoch. 
    
\begin{table}[]
\centering
\caption{Gender classification accuracy (\%) on the ND-GFI dataset with two classifiers: RDF and NNet.}
\label{tab:ndGfi}
\vspace{-5pt}
\begin{tabular}{|l|P{1cm}|P{1cm}|}
\hline
\textbf{Algorithm} & \textbf{RDF} & \textbf{NNet} \\
\hline
\hline
Stacked Autoencoder \cite{sae} & 64.17  & 65.33 \\
\hline
Stacked Denoising Autoencoder \cite{sdae} & 71.33 & 66.00 \\
\hline
Deep Belief Network \cite{dbn} & 67.17 & 71.83 \\
\hline
Discriminative RBM \cite{drbm} & \multicolumn{2}{c|}{78.67}  \\
\hline
\textbf{Proposed Deep Class-Encoder} & \textbf{78.17} & \textbf{83.17}\\
\hline
\end{tabular}
\vspace{-5pt}
\end{table}

\section{Conclusion}
This research addresses the task of gender and ethnicity classification of iris images. A supervised autoencoder model, termed as Deep Class-Encoder has been proposed for the given task. Deep Class-Encoder utilizes the class labels at the time of feature learning, in order to learn discriminative features. The efficacy of the proposed model is evaluated on two datasets each, for gender and ethnicity classification. Experimental evaluation and results further promote the utility of the proposed model for learning class-specific discriminative features. 

\section{Acknowledgment}
This research is partially supported by MEITY (Government of India), India. S. Nagpal is partially supported through TCS PhD fellowship. M. Vatsa, R. Singh, and A. Majumdar are partially supported through Infosys Center for Artificial Intelligence. We also thank NVIDIA Corp. for Tesla K40 GPU for research.

{\small
\bibliographystyle{ieee}
\bibliography{submission_example}
}

\end{document}